\documentclass[letterpaper]{article} 
\usepackage{aaai2026}  
\usepackage{times}  
\usepackage{helvet}  
\usepackage{courier}  
\usepackage[hyphens]{url}  
\usepackage{graphicx} 
\usepackage{booktabs} 
\usepackage{natbib}  
\usepackage{caption} 
\usepackage{algorithm}
\usepackage{algorithmic}

\usepackage{newfloat}
\usepackage{listings}
\DeclareCaptionStyle{ruled}{labelfont=normalfont,labelsep=colon,strut=off} 
\floatstyle{ruled}
\newfloat{listing}{tb}{lst}{}
\floatname{listing}{Listing}
\usepackage{amsmath}
\usepackage{multirow} 

\makeatletter
\def\@oddfoot{\hfil \thepage \hfil}  
\def\@evenfoot{\hfil \thepage \hfil} 
\makeatother

\nocopyright

\title{SafeCtrl: Region-Based Safety Control for Text-to-Image Diffusion via Detect-Then-Suppress}

\author {
    Lingyun Zhang\textsuperscript{\rm 1},
    Yu Xie\textsuperscript{\rm 2}\thanks{Corresponding author.},
    Yanwei Fu\textsuperscript{\rm 1},
    Ping Chen\textsuperscript{\rm 1}
}
\affiliations {
    \textsuperscript{\rm 1}Fudan University\\
    \textsuperscript{\rm 2}Purple Mountain Laboratories\\
    lyzhang22@m.fudan.edu.cn, xieyu9332@gmail.com, yanweifu@fudan.edu.cn,
    pchen@fudan.edu.cn 
}

\usepackage{bibentry}

\begin{document}

\maketitle

\begin{abstract}
The widespread deployment of text-to-image models is challenged by their potential to generate harmful content. While existing safety methods, such as prompt rewriting or model fine-tuning, provide valuable interventions, they often introduce a trade-off between safety and fidelity. Recent localization-based approaches have shown promise, yet their reliance on explicit ``concept replacement" can sometimes lead to semantic incongruity.
To address these limitations, we explore a more flexible detect-then-suppress paradigm. We introduce SafeCtrl, a lightweight, non-intrusive plugin that first precisely localizes unsafe content. Instead of performing a hard A-to-B substitution, SafeCtrl then suppresses the harmful semantics, allowing the generative process to naturally and coherently resolve into a safe, context-aware alternative.
A key aspect of our work is a novel training strategy using Direct Preference Optimization (DPO). We leverage readily available, image-level preference data to train our module, enabling it to learn nuanced suppression behaviors and perform region-guided interventions at inference without requiring costly, pixel-level annotations. Extensive experiments show that SafeCtrl significantly outperforms state-of-the-art methods in both safety efficacy and fidelity preservation. Our findings suggest that decoupled, suppression-based control is a highly effective and scalable direction for building more responsible generative models.
\end{abstract}

\section{Introduction}
Text-to-image diffusion models~\cite{sohl2015deep, ho2020denoising, song2021denoising, rombach2022high, saharia2022photorealistic} have revolutionized visual content generation, yet their widespread deployment is critically hampered by their propensity to generate harmful content, such as nudity, violence, or biased imagery. Unsafe Diffusion\cite{qu2023unsafe} shows that a non-negligible proportion of outputs from popular diffusion models (e.g., up to 18.92\% for Stable Diffusion) contain harmful elements. This challenge \cite{birhane2021multimodal} largely stems from unsafe data inherited from web-scale training corpora~\cite{schuhmann2022laionb, schuhmann2021laion}.

\begin{figure}[t]
\centering
\includegraphics[width=1.0\linewidth]{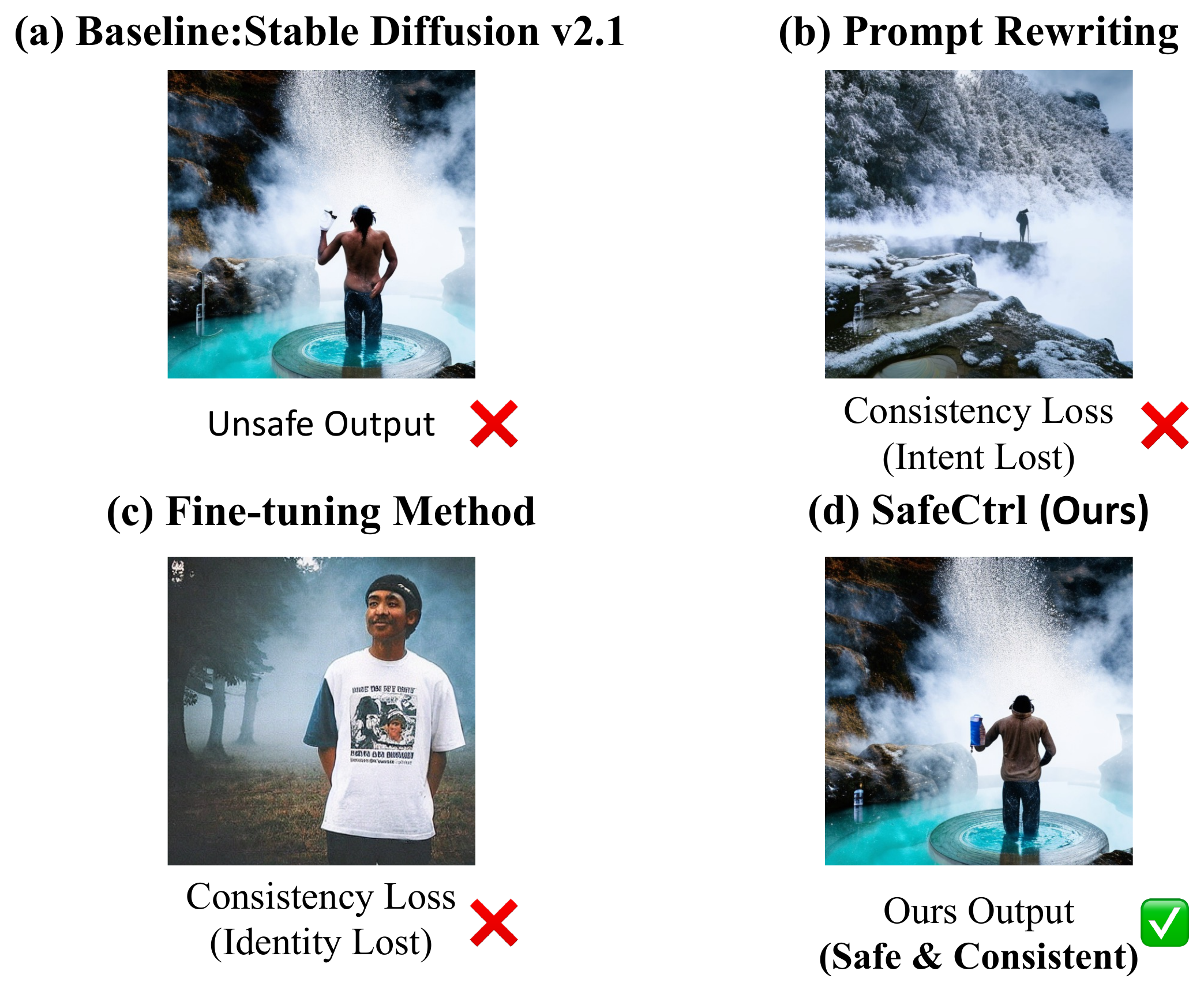} 
\caption{
    \textbf{Comparison of safety interventions on an artistic prompt.}
    The original prompt was \textit{``An artist rinsing off sweat in a snowy mountain hot spring, with mist swirling around"}.
    \textbf{(a)} The baseline generates \textbf{explicit content} despite capturing the artistic intent.
    \textbf{(b)} \textbf{Prompt Rewriting} adds overhead and rewrites the prompt to ``...taking pictures...fully clothed...", causing a catastrophic loss of the original scene and intent.
    \textbf{(c)} \textbf{Fine-tuning} fails to maintain the subject's identity.
    \textbf{(d)} In contrast, our non-intrusive \textbf{SafeCtrl} is the only method shown to produce a safe, consistent, and faithful output, demonstrating a superior solution to the safety-fidelity trade-off.
}
\label{fig:teaser}
\end{figure}

To address this, prevailing safety interventions have emerged, but they typically fall into two categories, both facing a fundamental dilemma. The first category, operating on the input/output level (eg, prompt rewriting\cite{yang2024guardt2i, wu2024universal} or rejection sampling~\cite{rando2022red}), acts as a brittle outer shell, often sacrificing the user's creative intent and being easily circumvented. The second, more robust category, intervenes at the model level (eg, fine-tuning ~\cite{gandikota2023erasing, schramowski2023safe, wu2025unlearning}). While more effective, these intrusive methods often result in a degradation of image quality or a loss of semantic coherence. This recurring trade-off, we believe, is a direct consequence of the tight coupling between the safety mechanism and the core generative process.
Recent modular approaches, like Concept Replacer (CR)~\cite{zhang2025concept}, have taken a promising step towards decoupling by localizing and replacing concepts. 
While effective, their ``hard replacement" paradigm can still introduce semantic artifacts and lacks adaptability. This suggests that a more elegant and effective paradigm may lie not in replacement, but in suppression.

Therefore, we introduce \textbf{SafeCtrl}, a framework that pioneers a detect-then-suppress paradigm for safety control. SafeCtrl operates as a lightweight, non-intrusive plugin that works in synergy with a frozen, pretrained diffusion model. It first \textit{detects} and localizes unsafe content using a novel unsafe attention module. Then, instead of performing a hard A-to-B substitution, it \textit{suppresses} the harmful semantics, allowing the generative process to naturally and coherently resolve into a safe alternative that respects the original context. This is achieved by dynamically modulating the input text embedding, effectively neutralizing the influence of unsafe concepts only in the localized regions.

A key innovation lies in our training and inference strategy for this nuanced control. While prior works like Safety-DPO~\cite{liu2024safetydpo} have applied DPO to globally fine-tune the entire model, this can inadvertently affect safe regions of an image. We introduce a novel approach: we leverage Direct Preference Optimization (DPO) on readily available, image-level preference data to train our lightweight, external control module. Crucially, the fine-grained, region-guided suppression is applied only at inference time. This unique decoupling of global training from local intervention allows SafeCtrl to learn nuanced safety behaviors without requiring costly, pixel-level annotations, and ensures that modifications are precisely targeted, leaving safe areas entirely untouched.

Our contributions are summarized as follows:
\begin{itemize}
\item We identify the ``tight coupling" of safety and generation as a core problem and propose a novel detect-then-suppress paradigm as a more flexible, decoupled solution.
\item We instantiate this paradigm in SafeCtrl, a lightweight, non-intrusive plugin featuring a novel unsafe attention module for real-time detection and a guidance mechanism for targeted suppression.
\item We propose a novel training strategy that leverages image-level DPO to learn region-guided control, eliminating the need for expensive pixel-level supervision.
\item We demonstrate through extensive experiments that SafeCtrl significantly outperforms a wide range of state-of-the-art methods in both safety efficacy and fidelity preservation.
\end{itemize}

\section{Related Work}

\subsection{Safety Control via Localization and Intervention}
Controlling harmful content generation\cite{somepalli2023diffusion} in diffusion models\cite{Rombach_2022_CVPR, ramesh2021zero} is a critical research area. Early approaches relied on brittle input/output filters~\cite{rando2022red}. A more robust line of work intervenes directly in the generative process. One major category involves \textbf{global model editing}, where methods like ESD~\cite{gandikota2023erasing}, UCE~\cite{gandikota2024unified}, and SLD~\cite{schramowski2023safe} attempt to ``erase" or steer away from unsafe concepts by fine-tuning the entire model, particular neurons\cite{yang2024pruning} or lora\cite{hu2022lora}. While often effective, these intrusive techniques risk degrading the model's general capabilities and content consistency.

Many studies\cite{chen2024artadapter, zhang2023adding, wang2023context, wu2023paragraph,mou2024t2i,ye2023ip} have focused on how to control model generation. To enable more precise interventions, another category focuses on \textbf{region-based control}.  These methods first localize a concept and then apply a targeted modification. In the context of general-purpose editing, works like DIFFEDIT~\cite{couairon2022diffedit} have shown the potential of using attention maps for localization. For safety, Concept Replacer (CR)~\cite{zhang2025concept} adopted this idea, using a heavyweight, duplicated U-Net\cite{ronneberger2015u} for segmentation and then performing a ``hard replacement" of the concept.

\textbf{Our approach} builds upon the insights of region-based control but introduces a fundamentally different paradigm. Instead of performing intrusive global edits or relying on a costly external segmentation model, SafeCtrl operates as a lightweight, non-intrusive plugin. More importantly, we shift the intervention paradigm from CR's ``hard replacement" to a more flexible ``semantic suppression". This allows the model to naturally generate a safe and coherent alternative, rather than being forced to insert a potentially incongruous concept.

\subsection{Preference Alignment for Generative Models}
Aligning generative models with human preferences, rather than relying on explicit rules, has become a dominant paradigm, popularized by RLHF~\cite{ouyang2022training} and more recently, Direct Preference Optimization (DPO)~\cite{rafailov2023direct}. DPO has been successfully applied to align diffusion models with global properties like aesthetics~\cite{wallace2024diffusion}. The most relevant work, Safety-DPO~\cite{liu2024safetydpo}, uses image-level preferences to globally fine-tune the entire model for safety.

\textbf{Our approach} proposes a novel application of DPO that bridges the gap between global supervision and local control. While Safety-DPO performs a global update that can inadvertently affect safe image regions, we are the first to use image-level DPO to train a decoupled, region-specific control module. Our framework learns a general safety alignment from holistic image preferences, while the precise, spatially-targeted suppression is applied only at inference time. This strategy bypasses the need for costly, pixel-level safety annotations, offering a more scalable and precise path to safety alignment.

\begin{figure*}[t]
\centering
\includegraphics[width=0.9\textwidth]{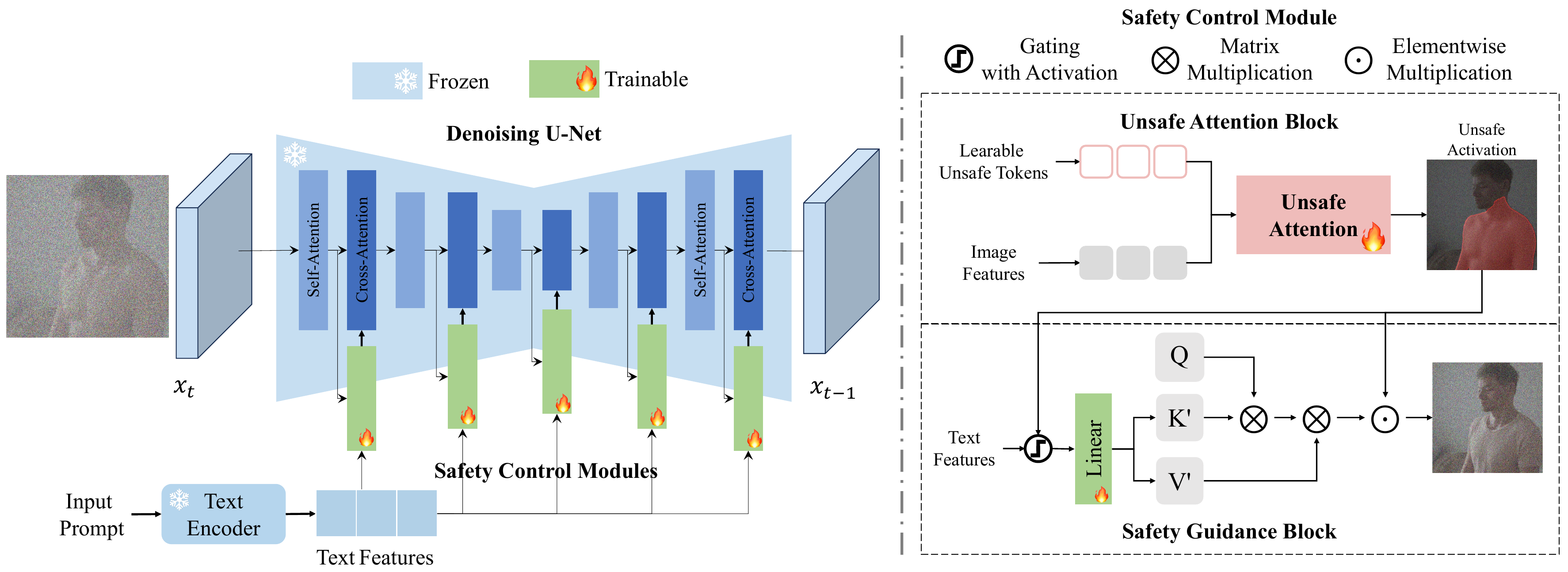} 
\caption{The overall architecture of SafeCtrl. SafeCtrl operates as a set of trainable, non-intrusive modules integrated into a frozen Denoising U-Net. At each step $t$, it receives the noisy latent $x_t$, text features, and the input prompt. 
The Safety Control Module is detailed in the Right. The \textit{Unsafe Attention Block} uses learnable tokens to detect potentially unsafe regions, producing a spatial mask. Concurrently, the \textit{Safety Guidance Block} processes text features and the localization information to generate a guidance signal, which modulates hidden states to steer the generation towards a safe and consistent output $x_{t-1}$.}
\label{fig:method_architecture}
\end{figure*}

\section{Method}
\subsection{Preliminaries: Text-to-Image Diffusion Models}
Text-to-image diffusion models\cite{dhariwal2021diffusion}, such as Stable Diffusion~\cite{rombach2022high}, have demonstrated remarkable capabilities in generating high-fidelity images from textual prompts. 
These models operate by iteratively denoising a latent variable \(z_t\) over a series of timesteps \(t = T, \dots, 1\), starting from pure Gaussian noise \(z_T \sim \mathcal{N}(0, I)\). 
At each step, a U-Net-based denoising network \(\epsilon_\theta(z_t, t, c)\) predicts the noise to be removed from \(z_t\), conditioned on the timestep \(t\) and an external context \(c\).

For text-to-image generation, the context \(c\) is derived from the input prompt via a text encoder, eg, CLIP~\cite{radford2021learning}. 
The semantic information from \(c\) is then injected into the U-Net's intermediate layers, typically through cross-attention mechanisms~\cite{vaswani2017attention}. 
This allows the model to align the visual generation with the textual description. 
Formally, the cross-attention output for a set of visual feature queries \(Q\) and text feature keys \(K\) and values \(V\) is computed as:
\begin{equation}
    \text{Attention}(Q, K, V) = \text{softmax}\left(\frac{QK^T}{\sqrt{d_k}}\right)V.
    \label{eq:cross_attention}
\end{equation}
where \( d_k \) is the dimension of queries.
Despite their success, the reliance on massive, often unfiltered web-scale datasets for training makes these models susceptible to generating harmful or unsafe content that is implicitly associated with certain concepts in the data. 
Our work aims to address this challenge by introducing an external safety control mechanism that operates on this very cross-attention process, which we detail in the following sections.

\subsection{The Detect-Then-Suppress Paradigm}
To address the safety challenges outlined in \S\ref{sec:preliminaries} without compromising the integrity of the pretrained model, we introduce a novel safety control framework named \textbf{SafeCtrl}. 
The core philosophy of our approach is a paradigm shift from direct concept removal or replacement to a more flexible, two-stage process: \textbf{Detect-Then-Suppress}. 
As illustrated in Figure~\ref{fig:method_architecture}, SafeCtrl operates as an external, lightweight control system that works in synergy with the frozen U-Net.
Our framework executes the following two stages at denoising:
\begin{enumerate}
    \item \textbf{Detection.} An \textit{Unsafe Attention Block} first analyzes the intermediate image features from the U-Net to identify and localize potentially harmful semantic regions. This produces a spatial risk map indicating ``what" to suppress and ``where".
    \item \textbf{Suppression.} A \textit{Safety Guidance Block} then takes this risk map and the original text features as input to compute a modified, ``safe" conditioning signal. This signal is then used to modulate the U-Net's cross-attention process, effectively steering the generation away from the unsafe concept only in the identified regions.
\end{enumerate}

This entire control mechanism is realized through a set of compact, trainable modules that are placed in parallel with the U-Net's attention layers, rather than being injected into them. 
This design ensures that our safety interventions are both precise and non-destructive.
The detailed implementations of the detection and suppression stages are presented in following sections.

\begin{table*}[t]
    \centering
    \begin{tabular}{c|c|cc|cc}
    \toprule
    \multirow{2}{*}{\textbf{Method}} & \textbf{Safety (on I2P)} & \multicolumn{2}{c|}{\textbf{Fidelity \& Alignment (on I2P)}} & \multicolumn{2}{c}{\textbf{Fidelity \& Alignment (on COCO-30k)}} \\ 
    \cmidrule(lr){2-2} \cmidrule(lr){3-4} \cmidrule(lr){5-6}
    & NudeNet Detections $\downarrow$ & FID $\downarrow$ & CLIP $\uparrow$ & FID $\downarrow$ & CLIP $\uparrow$ \\ 
    \midrule
    Base Model SD v2.1 & 586 & - & 31.65 & 14.87 & 31.53 \\
    \midrule
    FMN & 424 & 23.74 & 28.89 & 13.52 & 30.39 \\
    AC & 838 & 129.3 & 20.84 & 14.13 & 31.37 \\
    UCE & 182 & 14.82 & 31.32 & 16.09 & 31.29 \\
    SLD-M & 212 & 20.45 & 30.19 & 20.92 & 30.38 \\
    ESD-u & 180 & 21.93 & 29.62 & 14.11 & 30.34 \\
    MACE & 111 & 33.43 & 23.64 & \textbf{13.42} & 29.41 \\
    CR & 126 & 17.88 & 30.68 & 15.15 & 30.67 \\
    \textbf{Ours (SafeCtrl)} & \textbf{55} & \textbf{12.14} & \textbf{31.54} & 15.18 & \textbf{31.30} \\ 
    \bottomrule
    \end{tabular}
    \caption{
        \textbf{Main quantitative comparison of safety and fidelity.} 
        Our method, SafeCtrl, is evaluated against state-of-the-art baselines. We report safety via NudeNet detections on the I2P benchmark (lower is better), and fidelity/alignment via FID (lower is better) and CLIP Score (higher is better) on both I2P and COCO-30k. 
        The results show that SafeCtrl achieves the \textbf{best safety performance} (lowest detection count) while simultaneously maintaining fidelity scores that are the \textbf{closest to the original SD v2.1 baseline}, demonstrating a superior balance between the two competing objectives. A detailed breakdown of NudeNet detection categories is provided in the Appendix.
    }
    \label{tab:safety_fidelity}
\end{table*}

\subsection{Unsafe Region Detection}
The first stage of our framework is to accurately and efficiently localize potentially unsafe regions within the image as it is being generated. 
To achieve this, we introduce a lightweight \textbf{Unsafe Attention Block} that operates externally, in parallel with the frozen U-Net layers. 
Our key insight is that we can repurpose the rich, multi-level features from the U-Net as a powerful input to our specialized detection module, thus avoiding the need for a separate, heavyweight segmentation model like in CR~\cite{zhang2025concept}.

\paragraph{Spatial Risk Map Estimation.}
Our Unsafe Attention Block computes a spatial risk map by matching image features against a set of learnable ``unsafe concept" prototypes. 
Specifically, we introduce a small set of learnable tokens, \(\mathcal{T}_{\text{unsafe}}\), which are trained to act as general-purpose semantic embeddings for various unsafe concepts (eg, one token might learn to represent ``nudity", another ``violence").

At each denoising step \(t\), the block operates as follows:
First, the input image features from the U-Net are passed through a learnable projection head to generate the query tensor \(Q\). 
Simultaneously, our learnable unsafe tokens \(\mathcal{T}_{\text{unsafe}}\) are passed through a separate projection head to produce the key \(K\) and value \(V\) tensors. 

The core of our detection mechanism is then a cross-attention operation between the image queries and the unsafe concept keys. The resulting attention map, \(A_{\text{cross}}\), effectively represents the spatial distribution of unsafe semantics by measuring the alignment between each image region (\(Q\)) and the learned unsafe prototypes (\(K\)).
A self-attention map, \(A_{\text{self}}\), is also computed from the image features themselves to capture spatial context. 
These two maps are then fused to produce the final spatial risk mask \(M\), which robustly highlights the unsafe regions:
\begin{equation}
    M = \text{vec}(A_{\text{cross}}) \cdot A_{\text{self}}.
    \label{eq:spatial_risk}
\end{equation}
where \(\text{vec}(\cdot)\) denotes the operation of flattening the cross-attention map into a vector.

\paragraph{Training Objective.}
\label{par:training_objective}
Our training process is designed to be highly parameter-efficient. Instead of fine-tuning the entire U-Net, we only train the parameters within our Unsafe Attention Block: the set of learnable unsafe tokens \(\mathcal{T}_{\text{unsafe}}\) and the lightweight projection heads for \(Q\) and \(K\). 
This allows for rapid, few-shot adaptation to new concepts with minimal annotation and computational cost.

To effectively teach our module to identify unsafe regions, we employ a composite loss function guided by ground-truth masks \(M'\). The total loss \(\mathcal{L}_{\text{detect}}\) is a weighted sum of a cross-entropy loss and a Mean Squared Error (MSE) loss:
\begin{equation}
    \mathcal{L}_{\text{detect}} = \lambda_{\text{ce}} \underbrace{\text{CE}(A_{\text{cross}}, M')}_{\mathcal{L}_{\text{CE}}: \text{Semantic Alignment}} + \lambda_{\text{mse}} \underbrace{\| M - M' \|^2_2}_{\mathcal{L}_{\text{MSE}}: \text{Spatial Refinement}}.
    \label{eq:loss_detect}
\end{equation}
The cross-entropy term aligns our learnable tokens with the unsafe semantics in labeled regions, while the MSE term refines the final fused mask's spatial accuracy.

\paragraph{Temporal Smoothing at Inference.}
To ensure stable localization across the diffusion process and reduce noisy fluctuations between timesteps, we apply exponential smoothing to the predicted masks at inference time. The smoothed mask \(\bar{M}_t\) used for suppression is computed as:
\begin{equation}
    \bar{M}_t = \alpha \cdot \bar{M}_{t-1} + (1-\alpha) \cdot M_t,
    \label{eq:smoothing}
\end{equation}
where \(M_t\) is the raw risk mask at timestep \(t\), and we set \(\alpha=0.3\). This temporal consistency is crucial for effective and stable guidance in the suppression stage.

\subsection{Suppression via Preference-Aligned Guidance}

Once an unsafe region is identified, the \textbf{Safety Guidance Block} performs a targeted intervention to steer the generation towards a safe outcome. Our core idea is to modulate the cross-attention mechanism, which is responsible for injecting textual semantics into the image features. Instead of altering the U-Net's architecture, we dynamically compute a ``safe" conditioning signal and apply it only within the detected unsafe regions, ensuring a minimally invasive yet effective suppression.

\begin{figure*}[t]
\centering
\includegraphics[width=0.9\textwidth]{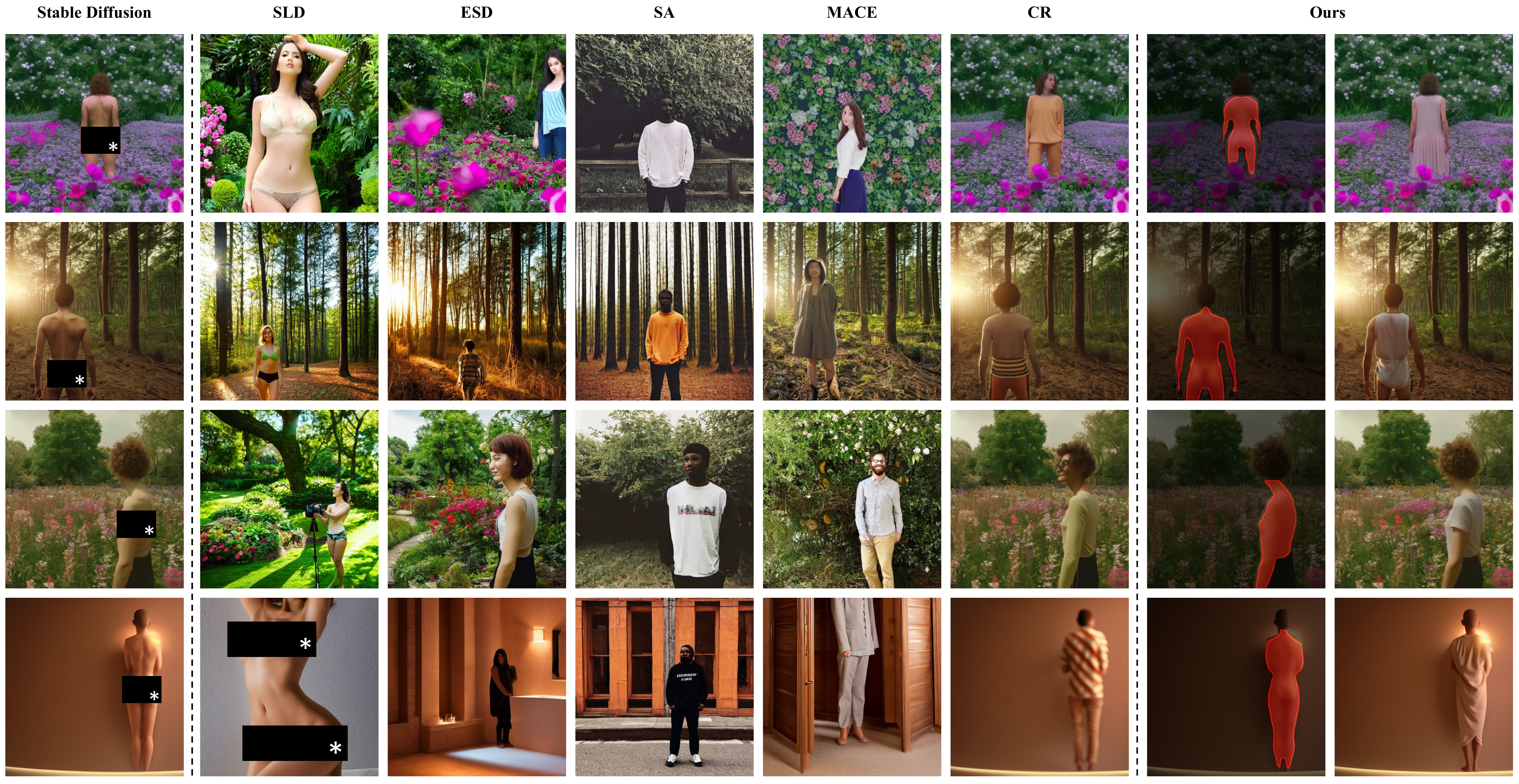} 
\caption{Qualitative comparison with state-of-the-art safety methods for nudity suppression. For each prompt, SafeCtrl is the only method that effectively removes the unsafe content while preserving high consistency with the original model's identity, background, and artistic style. In contrast, prior methods either fail to ensure safety or suffer from severe consistency loss and quality degradation.}
\label{fig:nudity_suppress}
\end{figure*}

\paragraph{Inference-Time Suppression Mechanism.}
In a standard cross-attention layer (Eq.~\eqref{eq:cross_attention}), the image features \(f\) attend to the text features \(c\) to produce an output \(Z\). Our guidance block introduces a parallel path. It first computes a set of ``safe" text features by passing the original text embeddings \(c\) through a learnable projection layers, \(W_K'\) and \(W_V'\), to get \(K'\) and \(V'\). These are used to compute a parallel, safety-guided attention output \(Z'\):
\begin{equation}
    Z' = \text{Softmax}\!\Bigl(\tfrac{QK'^\top}{\sqrt d}\Bigr)\,V', 
    \label{eq:safe_attention}
\end{equation}
\text{where } \(Q = W_Q f \), \(K' = W_K' c\), \(V' = W_V' c\).
Finally, we fuse the original attention output \(Z\) and our safety-guided output \(Z'\) using the smoothed risk map \(\bar{M}\) from the detection stage. This fusion is performed via element-wise blending, confining the intervention only to the unsafe areas:
\begin{equation}
    Z_{\text{new}} = Z \odot (1 - \bar{M}) \;+\; Z' \odot \bar{M}.
    \label{eq:blending}
\end{equation}
The resulting feature map \(Z_{\text{new}}\) is then passed to the next layer. This design ensures that the safety guidance is applied precisely where needed, preserving the original model's output in all safe regions.

\paragraph{Training with Preference Alignment via DPO.}
We train the lightweight projection layers \(W_K'\) and \(W_V'\) of the guidance block using Direct Preference Optimization (DPO)~\cite{rafailov2023direct}. Our key innovation is to leverage readily available, \textbf{image-level} preference pairs (\(y_w, y_l\)), where \(y_w\) is a preferred (safe) image and \(y_l\) is a rejected (unsafe) image. The DPO loss \(\mathcal{L}_{\text{DPO}}\) trains our SafeCtrl-equipped model \(\epsilon_\theta\) to better denoise preferred examples and worse denoise rejected ones, compared to a frozen reference model \(\epsilon_{\text{ref}}\) (see Appendix for the full loss equation).

This strategy is highly efficient as it decouples training from localization: the module learns a general-purpose safety alignment from global image preferences, while the precise, region-guided suppression (Eq.~\eqref{eq:blending}) is applied only during inference. This avoids the need for costly, pixel-level safety annotations during preference tuning.

\section{Experiments}
We conduct a comprehensive set of experiments to rigorously evaluate our proposed SafeCtrl framework. Our evaluation is designed to demonstrate SafeCtrl's superiority in navigating the critical trade-off between Safety and Fidelity. To this end, we first present our main end-to-end results, then analyze the core detection module, and conclude with extensive ablation studies.

\subsection{Experimental Setup}
We implement SafeCtrl on top of the frozen Stable Diffusion v2.1~\cite{rombach2022high}. Our evaluation is conducted on several standard benchmarks. \textbf{For localization accuracy}, we train our detection module in a few-shot setting (10 images per concept) and evaluate on CelebA-Mask-HQ, Pascal-Car, and Pascal-Horse test splits using mean Intersection over Union (mIoU). \textbf{For safety suppression}, we evaluate on the I2P benchmark~\cite{schramowski2023safe} by measuring the detection rate of the NudeNet~\cite{bedapudi2019nudenet} classifier. \textbf{For fidelity and alignment}, we report Fréchet Inception Distance (FID)~\cite{heusel2017gans} and CLIP Score~\cite{radford2021learning} on both I2P and COCO-30k~\cite{lin2014microsoft}. We compare SafeCtrl against a comprehensive set of state-of-the-art methods, including SLD, ESD, CR, and others. \textbf{All implementation details, including dataset construction, hyperparameters, and the full list of baselines, are provided in the Appendix.}

\subsection{Main Results: Safety and Fidelity Comparison}
We now present our main results, evaluating SafeCtrl against a comprehensive suite of state-of-the-art safety methods on the critical trade-off between safety and fidelity. Baselines include global fine-tuning methods (ESD, UCE, SA, MACE), guidance-based approaches (SLD), and Concept Replacer (CR). A detailed description of each baseline is provided in the Appendix.

\begin{figure}[t]
\centering
\includegraphics[width=0.7\linewidth]{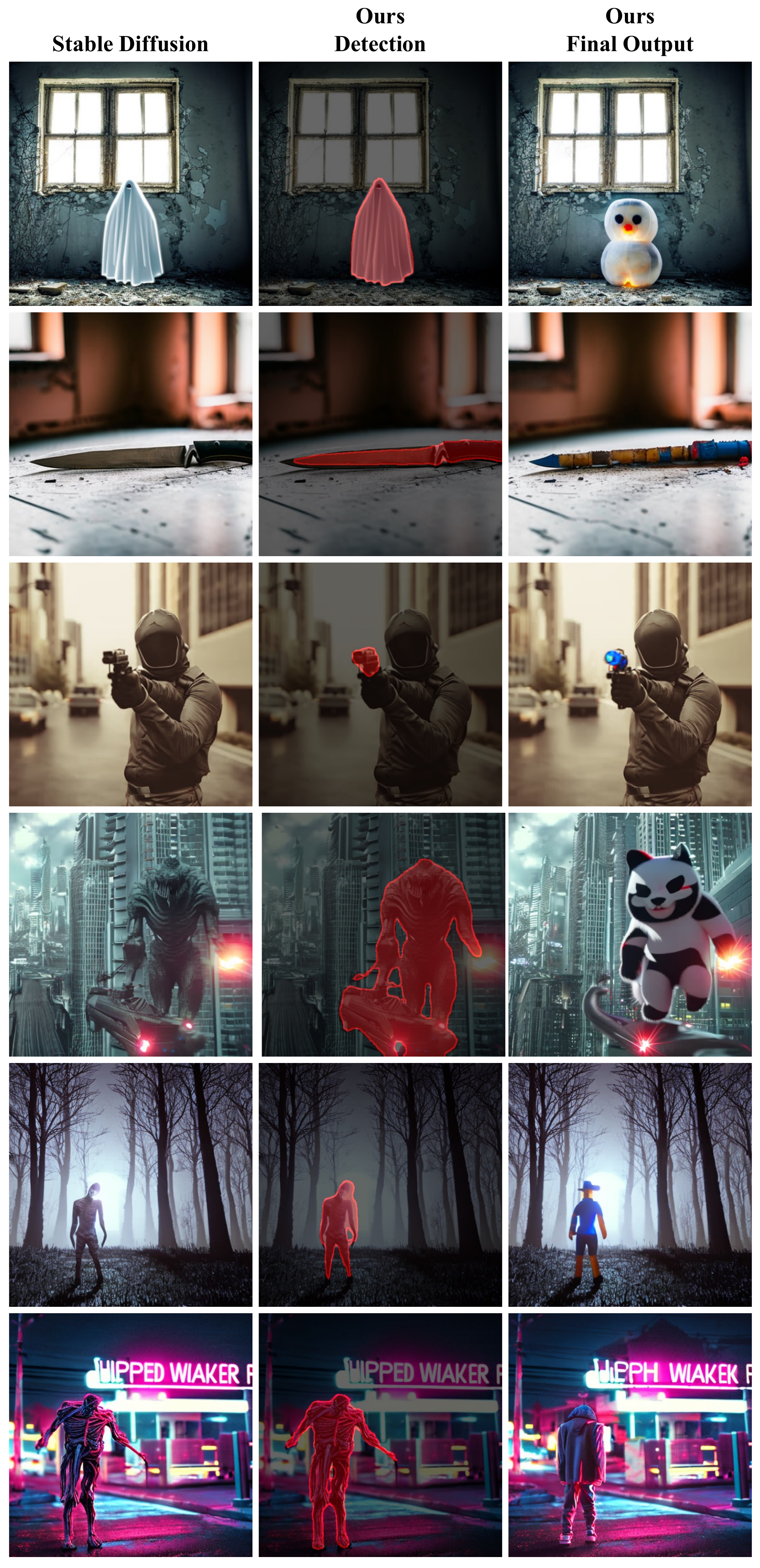} 
\caption{Qualitative results to other sensitive concepts. The final outputs highlight our method's ability to perform context-aware and plausible concept modification. By suppressing the unsafe semantics, SafeCtrl guides the generation towards coherent and creative alternatives (e.g., a ghost pivoting to a snowman, a monster concept being steered towards a cat, or a gun transforming into a toy).}
\label{fig:multiconcept_suppress}
\end{figure}

\paragraph{Quantitative Analysis.}
Table~\ref{tab:safety_fidelity} summarizes our quantitative results. 
\textbf{For safety}, we evaluate on the I2P benchmark, measuring the number of unsafe images generated (lower is better). SafeCtrl achieves the lowest detection count by a significant margin, demonstrating its superior suppression capabilities compared to all baselines. 
\textbf{For fidelity and alignment}, we report FID and CLIP scores on both I2P and the general-purpose COCO-30k benchmark. 
As shown, SafeCtrl consistently maintains scores that are closest to the original Stable Diffusion v2.1, indicating minimal impact on the model's generative distribution. 
This quantitatively validates that our method ensures safety without sacrificing image quality or semantic coherence.

\paragraph{Qualitative Analysis.}
Figure~\ref{fig:nudity_suppress} provides a direct qualitative comparison. The visual evidence corroborates our quantitative findings and reveals the distinct failure modes of prior methods. 
For instance, SLD often fails to suppress nudity, while global editing methods like ESD and SA suffer from severe consistency loss, drastically altering the subject's identity. 
Even the stronger region-based method, CR, while preserving identity, can introduce semantically incoherent elements (e.g., pajamas in a garden).
In contrast, \textbf{SafeCtrl} is the only method that consistently produces outputs that are both safe and highly faithful to the original prompt's identity, background, and artistic style. 
This visual superiority underscores the effectiveness of our detect-then-suppress paradigm.

\paragraph{Generalization to Other Concepts.}
Beyond nudity, SafeCtrl demonstrates strong generalization to a wider array of sensitive concepts, including horror themes (ghosts, monsters) and dangerous objects (knives, guns) in Figure~\ref{fig:multiconcept_suppress}. It successfully localizes and suppresses these varied concepts while maintaining contextual coherence. Detailed qualitative results for these concepts are presented in the Appendix.

\begin{table}[t]
    \centering
    \setlength{\tabcolsep}{1mm} 
    \begin{tabular}{l|c|c|c} 
    \toprule
    \textbf{Method} & \textbf{Pascal-Car} & \textbf{CelebA-HQ} & \textbf{Pascal-Horse} \\
    \midrule
    \multicolumn{4}{l}{\textit{Supervised Baselines}} \\
    CNN+CRF & 57.0 & - & - \\
    SegDDPM & - & 78.0 & 58.7 \\
    \midrule
    \multicolumn{4}{l}{\textit{Few-Shot / Unsupervised Methods}} \\
    ReGAN & 52.2 & 69.9 & 54.3 \\
    SLiMe & 68.3 & 75.7 & 63.3 \\
    CR & 69.3 & 78.1 & 64.3 \\
    \midrule
    \textbf{Ours (SafeCtrl)} & \textbf{72.1} & \textbf{78.3} & \textbf{65.7} \\
    \bottomrule
    \end{tabular}
    \caption{
        \textbf{Quantitative evaluation of localization accuracy (mIoU $\uparrow$).}
        We evaluate our detection module on three standard benchmarks under a 10-shot setting: Pascal-Car, CelebA-Mask-HQ, and Pascal-Pascal-Horse.
        Our method consistently outperforms state-of-the-art unsupervised and few-shot localization approaches.
        Detailed per-category results are in the Appendix.
    }
    \label{tab:localization_results}
\end{table}

\begin{figure*}[t]
    \centering
    \includegraphics[width=0.9\textwidth]{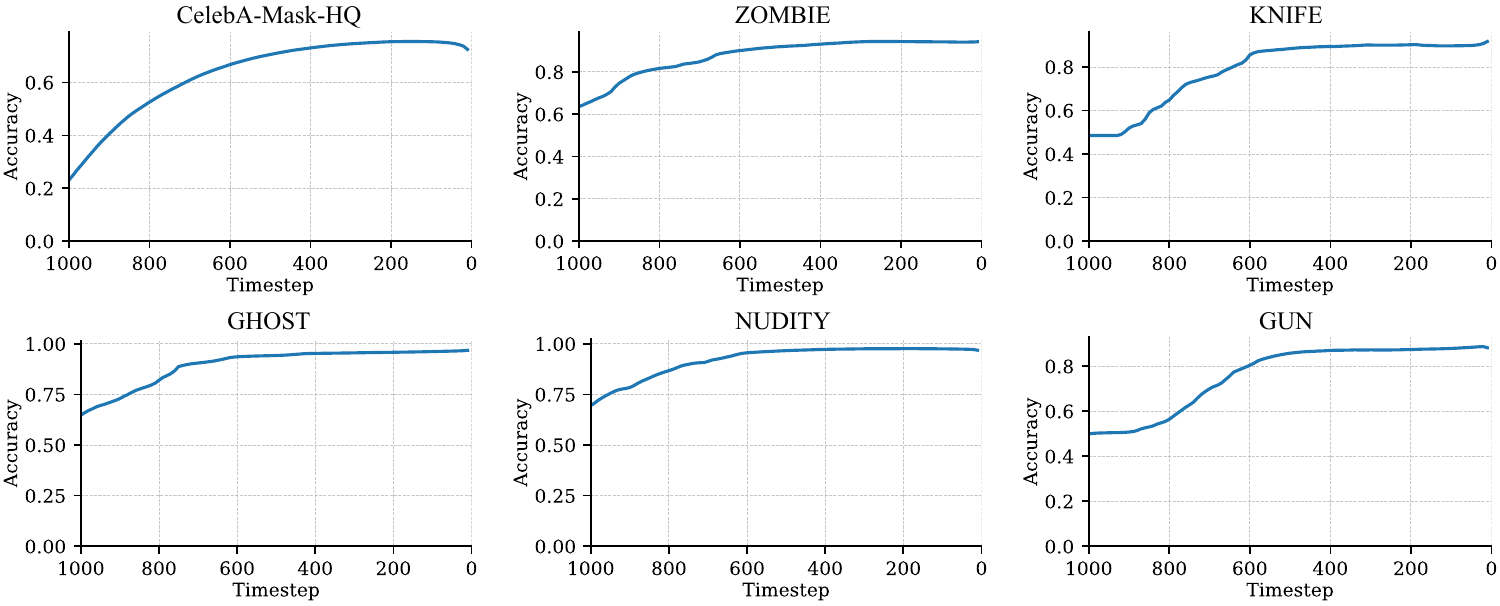} 
    \caption{
        \textbf{Impact of denoising timestep on localization accuracy.} 
        The mIoU for various concepts is plotted against the denoising timestep \(t\). Accuracy consistently improves as the timestep \(t\) decreases, confirming that semantic concepts become more identifiable in the later, less noisy stages of generation.
    }
    \label{fig:ablation_timestep}
\end{figure*}

\subsection{Analysis of the Detection Module}
To validate the effectiveness of our Unsafe Attention Block, we evaluate its localization accuracy both quantitatively and qualitatively. Since large-scale, pixel-annotated unsafe datasets are unavailable, we follow prior works~\cite{zhang2025concept, khani2023slime, simsar2025lime} and assess performance on proxy segmentation tasks using standard benchmarks.

\paragraph{Quantitative Accuracy.}
We compare SafeCtrl against state-of-the-art unsupervised and few-shot localization methods, including ReGAN\cite{tritrong2021repurposing}, SLiMe\cite{khani2023slime}, and the most relevant baseline, CR. All methods are trained in a consistent 10-shot setting for a fair comparison. The detailed setup and a full list of baselines are in the Appendix.
As shown in Table~\ref{tab:localization_results}, which reports the mean IoU, \textbf{SafeCtrl consistently outperforms all competing methods across all three benchmarks}: Pascal-Car, CelebA-Mask-HQ, and Pascal-Horse. This strong quantitative performance is particularly noteworthy given that our method achieves this with a lightweight, in contrast to CR which requires a heavyweight, duplicated U-Net. This validates our core hypothesis: repurposing the U-Net's rich features is a highly effective and parameter-efficient strategy for localization.

\paragraph{Qualitative Visualization.}
Fig~\ref{fig:nudity_suppress} and Fig~\ref{fig:multiconcept_suppress} provides qualitative evidence of our module's precision and generalization on actual sensitive concepts. The visualizations demonstrate that SafeCtrl can accurately localize a wide variety of targets, from large, deformable subjects like `nudity' and `monsters', to small, fine-grained objects like `weapons'. This robust localization capability across diverse and challenging scenarios is the critical foundation that enables the success of our subsequent suppression stage. More examples are available in the Appendix.

\subsection{Ablation Studies}

We conduct a series of ablation studies to validate our key design choices. We first analyze the impact of the denoising timestep on detection accuracy, and then verify the effectiveness of our region-guided suppression mechanism.

\paragraph{Analysis of Detection Timestep.}
To justify our strategy of applying suppression primarily in the later stages of denoising, we first investigate at which timesteps unsafe concepts become semantically distinct. We evaluate the localization accuracy (mIoU) of our detection module across the entire denoising process (\(t=1000 \to 1\)) on various concepts. As illustrated in Figure~\ref{fig:ablation_timestep}, for all concepts, the accuracy remains low during the initial noisy stages but rises sharply and stabilizes in the later stages (e.g., \(t < 200\)), where coherent image structures emerge. This result empirically validates our design choice to activate suppression in the later, more semantically stable part of the generation, ensuring that our interventions are based on reliable detections.

\paragraph{Effectiveness of Region-Guided Suppression.}
To isolate the benefit of our spatially-targeted intervention, we compare our full SafeCtrl model against a strong baseline variant: `` w/o Region-Guidance". This variant uses the exact same DPO training on our plugin, but at inference time, the suppression is applied \textit{globally} to the entire image, without using the spatial mask from our detection module. 
Table~\ref{tab:ablation_dpo} shows the comparison. While the global suppression approach achieves a slightly lower NudeNet detection count, this comes at a significant cost to image fidelity, evidenced by its much worse FID and CLIP scores. Our full, region-guided SafeCtrl, in contrast, achieves a superior balance. It maintains high fidelity (FID 12.14, CLIP 31.54) by applying modifications only where necessary. This clearly demonstrates that our region-guided suppression is the critical component for preserving content consistency and avoiding the collateral damage often seen in global safety methods.

\begin{table}[h]
    \centering
    \setlength{\tabcolsep}{0.5mm}
    \begin{tabular}{c|ccc}
    \toprule
    \textbf{Method} & NudeNet $\downarrow$ & FID (I2P) $\downarrow$ & CLIP (I2P) $\uparrow$ \\ 
    \midrule
    w/o Region-Guidance & \textbf{48} & 20.69 & 30.53 \\
    \textbf{Ours (Full Model)} & 55 & \textbf{12.14} & \textbf{31.54}  \\
    \bottomrule
    \end{tabular}
    \caption{
        \textbf{Ablation study on region-guided suppression.} 
        While global suppression is slightly more aggressive on safety metrics, our region-guided approach achieves a significantly better trade-off, preserving much higher image fidelity and alignment.
    }
    \label{tab:ablation_dpo}
\end{table}

\section{Conclusion}
In this work, we introduced SafeCtrl, a novel framework designed to resolve the fundamental safety-fidelity trade-off in diffusion models. We identified the tight coupling of safety and generation in prior methods as the core problem and, in response, proposed a more flexible, decoupled detect-then-suppress paradigm. Our key innovation is a novel training strategy that leverages image-level Direct Preference Optimization to effectively train a region-specific control module, enabling precise, localized interventions without requiring costly pixel-level supervision. Extensive experiments validate that our lightweight plugin significantly outperforms state-of-the-art methods in achieving a superior balance between safety and fidelity. We believe our approach represents a significant and scalable step towards more controllable, interpretable, and trustworthy generative AI.
\bibliography{aaai2026}

\end{document}